\documentclass[10pt,twocolumn]{article}

\usepackage{iccv}
\usepackage{times}
\usepackage{epsfig}
\usepackage{graphicx}
\usepackage{amsmath}
\usepackage{amssymb}

\usepackage{booktabs}
\usepackage{cite}
\usepackage{siunitx}
\usepackage{multirow}
\usepackage{alias2}
\usepackage{epsfig}
\usepackage{graphicx}
\usepackage{balance}
\usepackage{rotating}
\usepackage[frozencache,cachedir=.]{minted}
\usepackage{xspace}
\usepackage[labelfont=bf]{caption}
\setminted{fontsize=\small,baselinestretch=1}

\usepackage[pagebackref=true,breaklinks=true,letterpaper=true,colorlinks,citecolor=blue,bookmarks=false]{hyperref}

\iccvfinalcopy 

\begin{document}

\urlstyle{same}
\title{PyRobot: An Open-source Robotics Framework for \\ Research and Benchmarking}

\author{Adithyavairavan Murali$^{*}$ \quad Tao Chen$^{*}$ \quad Kalyan Vasudev Alwala$^{*}$ \quad Dhiraj Gandhi$^{*}$\\ Lerrel Pinto \quad Saurabh Gupta \quad Abhinav Gupta \\
\\
Facebook AI Research \quad Carnegie Mellon University \\
\url{https://www.pyrobot.org}}

\maketitle

\blfootnote{$^*$The first four authors contributed equally to this paper.}
\begin{abstract}
This paper introduces \pyrobot, an open-source robotics framework for research and benchmarking. \pyrobot is a light-weight, high-level interface on top of ROS that provides a consistent set of hardware independent mid-level APIs to control different robots. \pyrobot abstracts away details about low-level controllers and inter-process communication, and allows non-robotics researchers (ML, CV researchers) to focus on building high-level AI applications. \pyrobot aims to provide a research ecosystem with convenient access to robotics datasets, algorithm implementations and models that can be used to quickly create a state-of-the-art baseline. We believe \pyrobot, when paired up with low-cost robot platforms such as \locobot, will reduce the entry barrier into robotics, and democratize robotics. \pyrobot is open-source, and can be accessed via \href{https://pyrobot.org}{https://pyrobot.org}.
\end{abstract}
\section{Introduction}
Over the last few years there have been significant advances in AI, specifically in the fields of machine learning, computer vision, natural language processing and speech. Most of these advancements have been fueled by high-capacity neural networks and the availability of large-scale datasets. However, an often overlooked reason for this fast-paced progress has been the development of a conducive research ecosystem. Platforms such as Caffe~\cite{jia2014caffe}, PyTorch~\cite{paszke2017automatic}, TensorFlow~\cite{tensorflow2015-whitepaper} have reduced the entry barrier, which has democratized and accelerated research in these fields. 
For example, a new researcher in computer vision can get started with training state-of-the-art detectors using PyTorch and MSCOCO~\cite{lin2014microsoft} in less than a day. Common platforms and datasets have also led to standardized evaluations and benchmarks which also helps quantify progress in these areas.

The field of data-driven robotics has also seen tremendous excitement and energy in the past several years~\cite{pinto2016supersizing, levine2016end, levine2018learning, mahler2017dex, anymal2019, openaihand, zhu2018dexterous, pinto2016curious, agrawal2016learning, finn2017deep, pinto2017supervision, robotsinhome}. However, compared to other areas in AI, it has been relatively hard for a new researcher to get started and contribute to the progress in robotics. Why is that the case? One obvious reason is that researchers have to set up significant hardware infrastructure. This creates a high entry-barrier for researchers both in terms of financial cost and development time. Fortunately, there has been substantial progress on this front with the development of low-cost robots such as Blue~\cite{blue2019}, LoCoBot~\cite{robotsinhome} and others~\cite{robotis, yang2019replab}. In fact, the cost of a robot is now comparable to that of the cost of a GPU! However even with these low-cost robots, getting started in robotics is still hard due to the lack of research platforms and a self-sustaining ecosystem. 

Frameworks such as ROS~\cite{quigley2009ros} have made setting up robots substantially easier by providing a common mid-level communication layer and tools that are agnostic to low-level hardware and program context. However, there are two issues with such open-source frameworks:

{\bf ROS requires expertise:} Dominant robotic software packages like ROS and MoveIt! are complex and require a substantial breadth of knowledge to understand the full stack of planners, kinematics libraries and low-level controllers. On the other hand, most new users do not have the necessary expertise or time to acquire a thorough understanding of the software stack. A light weight, high-level interface would ease the learning curve for AI practitioners, students and hobbyists interested in getting started in robotics.

{\bf Lack of hardware-independent APIs:} Writing hardware-independant software is extremely challenging. In the ROS ecosystem, this was partly handled by encapsulating hardware-specific details in the Universal Robot Description Format (URDF) which other downstream services could read from. Yet, from the perspective of high-level AI applications, most robotics code is still hardware dependent. As a community, we lack a research platform and a common API that we can use to share code, datasets and models.

In this white-paper, we attempt to tackle these challenges via an open-source research platform -- {\bf PyRobot}. \pyrobot is a light weight, high-level interface on top of ROS that provides hardware independent mid-level APIs and high-level examples for manipulation and navigation. \pyrobot also provides libraries for hand-eye calibration, tele-operation, trajectory tracking, and SLAM-based navigation.
We believe \pyrobot combined with the recently released LoCoBot robot will reduce both the financial cost and development time -- leading to democratization of data-driven robotics. The hardware-independent API will lead to development of code and datasets that can be shared across the community. While the current \pyrobot release interfaces with LoCoBot and Sawyer, we plan to release integration with several new robots like the UR5~\cite{ur5spec} and Franka~\cite{franka}, and simulator platforms like MuJoCo~\cite{todorov2012mujoco} and Habitat~\cite{habitat19arxiv}.
\section{PyRobot Framework}

\begin{figure}
  \begin{center}
    \includegraphics[width=\linewidth]{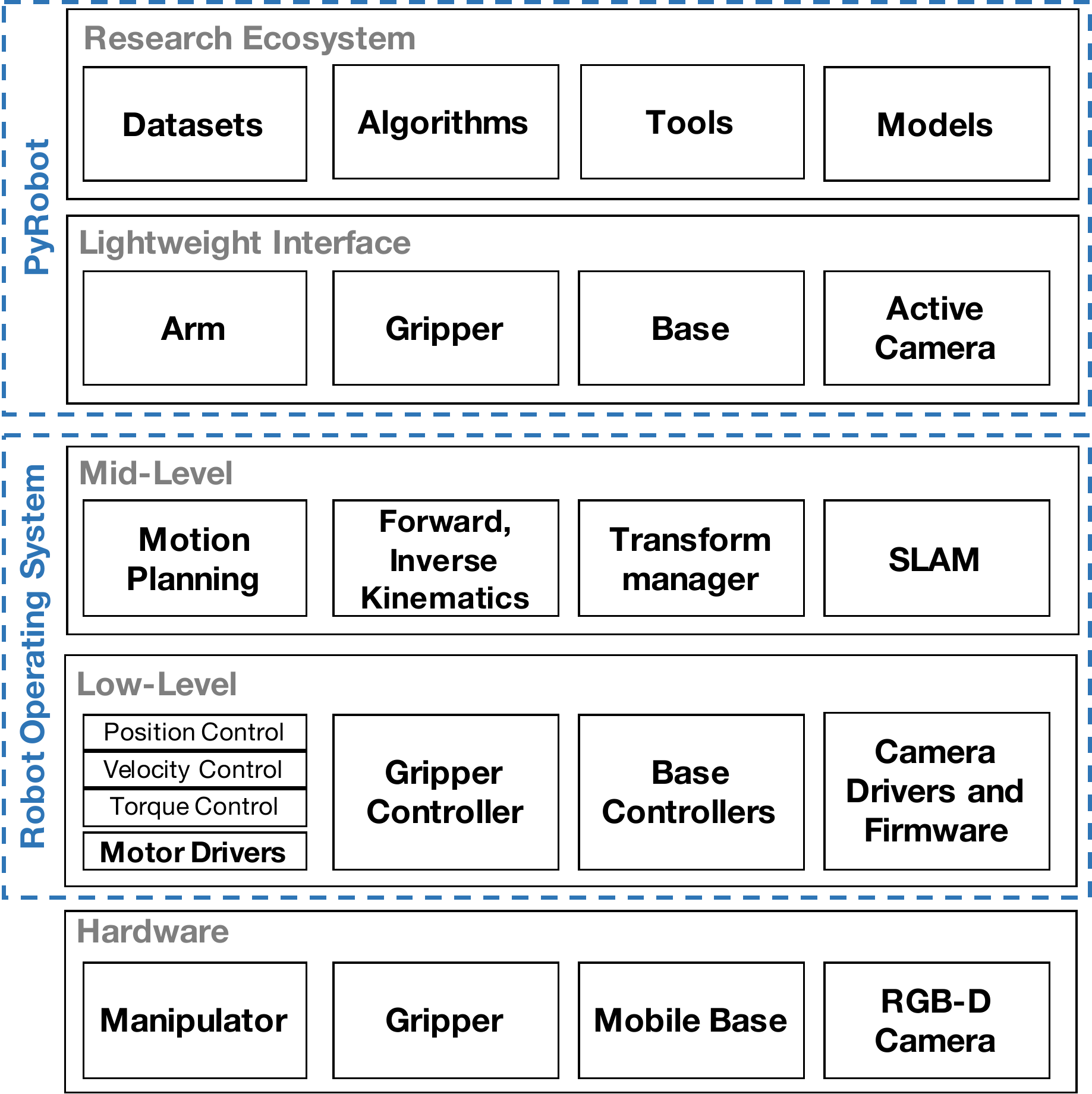}
    \caption{Overview of PyRobot system architecture.
}
  \label{fg:architecture}
  \end{center}
\end{figure}

PyRobot is a python-based robotics framework that isolates the ROS system~\cite{quigley2009ros} from the user-end and supports the same API across different robots (see Figure \ref{fg:architecture} for an overview). Essentially, it provides a python wrapper around the mid-level features provided by ROS and the low-level C++/C controllers and driver backends. PyRobot has common utility functions for all robots, such as joint position control, joint velocity control, joint torque control, cartesian path planning, forward kinematics and inverse kinematics (based on the robot URDF file), path planning, visual SLAM, among other features. Though it abstracts away the complexity of the underlying software stack, users still have the flexibility to use components at varying levels of the hierarchy, such as commanding low-level velocities and torques by-passing a planner. We summarize the design philosophy behind PyRobot below.


\begin{listing}[t]
\caption{PyRobot example for position control on LoCoBot and Sawyer.}
\label{code:jp_control}
\begin{minted}{python}
# LoCoBot - Arm
from pyrobot import Robot
bot = Robot('locobot')
target_joints = [0, 0, 0, 0, 0]
bot.arm.set_joint_positions(target_joints)

# LoCoBot - Base
target_position = [1, 1, 1]
bot.base.go_to_absolute(target_position)

# Sawyer
from pyrobot import Robot
bot = Robot('sawyer',
            use_arm=True,
            use_base=False,
            use_camera=False,
            use_gripper=True)
target_joints = [0, 0, 0, 0, 0, 0, 0]
bot.arm.set_joint_positions(target_joints)
\end{minted}
\end{listing}

\textbf{Beginner-friendly.} Ideally, new users should be able to start commanding a robot in just a few lines of code, as shown in the Listing~\ref{code:jp_control}, without learning ROS or the underlying software and firmware stack.

\textbf{Hardware-agnostic design.} PyRobot is designed to easily accommodate common robotic manipulators and mobile bases. Currently, it supports LoCoBot, a low-cost mobile robot with a 5-DOF manipulator and a Sawyer robot. Each robot has a YACS~\cite{yacs} configuration file that specifies the necessary robot-specific parameters: joint names, ROS topics to get state and set commands, base frame, end-effector frame, planner configuration, inverse kinematics solution tolerance, whether it has an arm or base or camera, \etc. A PyRobot object requires the config file for initialization. As shown in Listing~\ref{code:jp_control}, the Sawyer robot can be commanded in a manner identical to that of LoCoBot.

\textbf{Open Source.} Robotics systems development has typically been constrained to robotics experts in academia and industry with access to expensive and niche robotics systems. However, the extensive scope of artificial intelligence requires strong collaboration between researchers to build and maintain these large systems and one can contribute to all layers of the stack with open sourcing. Apart from the open software, LoCoBot works as an affordable open hardware that can be easily assembled for use with PyRobot. While simulation is useful for software testing and running experiments, writing software that works on the real robot is the eventual goal of the field and has severe challenges. As more developers have access to both open hardware and software, high quality applications tested on real robots can be publicly shared.

\section{Supported Hardware and Simulators}
\pyrobot is currently integrated with the following robots. In addition to real robots, \pyrobot can also be used to control robots in simulators like Gazebo.

\begin{figure}[t]
  \begin{center}
    \includegraphics[width=1.00\linewidth]{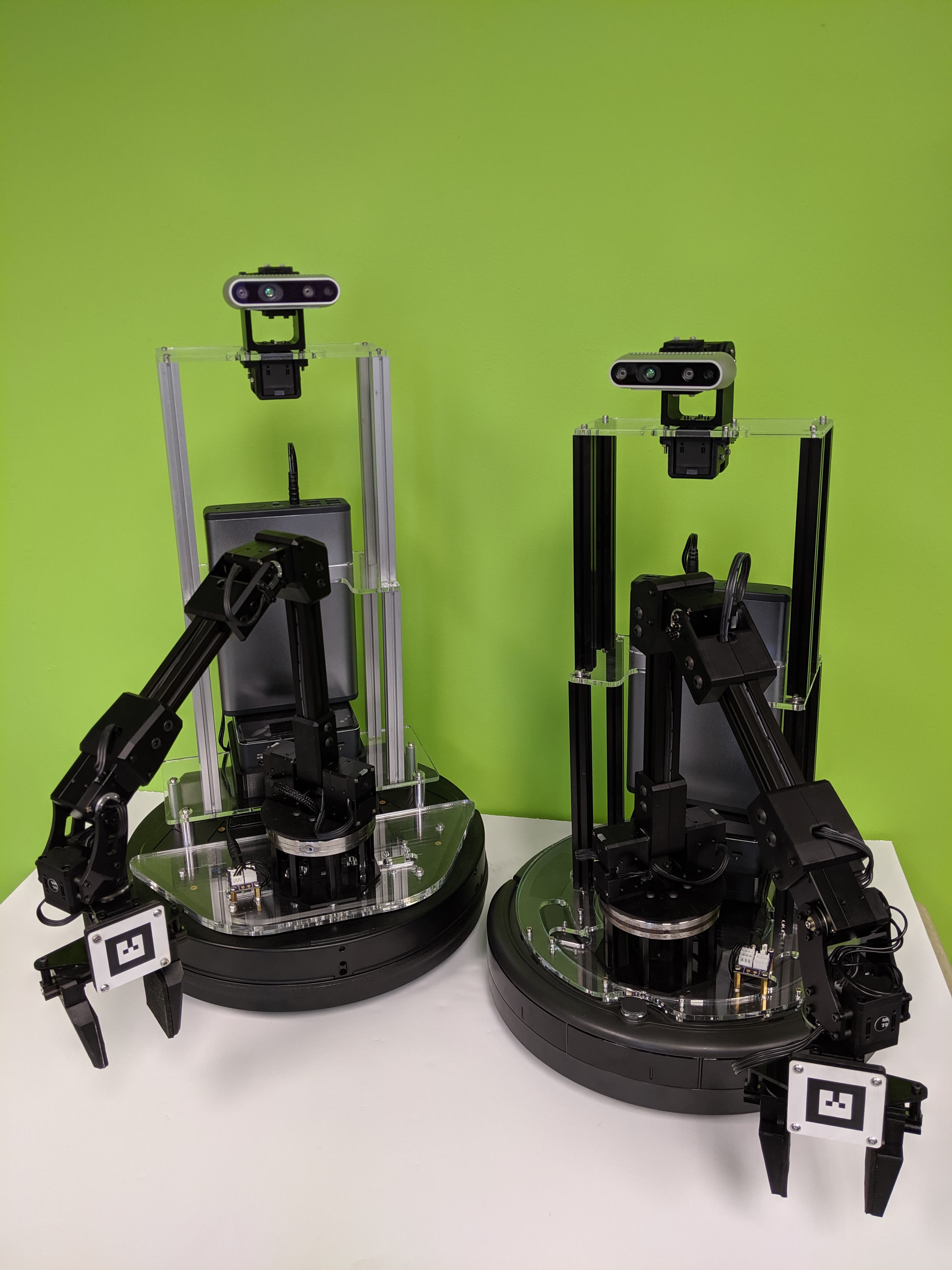} 
    \caption{\locobot (left) and \locobotlite (right). Both robots have a 5 DOF arm mounted on top of a mobile base (Kobuki or Create2). Robots are equipped with a RGB-D camera mounted on a pan-tilt stand. Robots come with a battery pack and an on-board computer.}
  \label{fg:cover_figure}
  \end{center}
\end{figure}

\textbf{LoCoBot}: LoCoBot, shown in Figure~\ref{fg:cover_figure}~(left), is a low-cost mobile manipulator platform built for easy setup and benchmarking robot learning research. It consists of a Trossen Widow X robotic arm~\cite{arm} assembled with Dynamixel XM-430 and XL-430s servo motors. The arm has five degrees of freedom (DOFs) - with a working payload of \SI{0.2}{\kg} and a maximum reach of \SI{0.55}{\m}. The robot comes in two versions, with the arm rigidly mounted on a Kobuki mobile base \cite{kobuki}. The Kobuki base is about \SI{0.12}{\m} high with payload capacity of around \SI{4.5}{\kg}. For visual perception, an Intel Realsense D435 RGBD camera \cite{camera} is mounted with a pan-tilt attachment at a height of about \SI{0.6}{\m} above the ground. An automatic camera calibration routine is implemented in the software suite. LoCoBot also comes with a Intel NUC (i5, 8GB RAM) machine rigidly attached on the base, which could be used for on-board compute. Kobuki base is powered through its own battery that can run base for about 2 hours. We use a 185 Wh battery pack \cite{battery} to power the arm, pan-tilt mount, and the on-board computer. On a full charge, the complete system is able to run for 50-60 minutes. \locobotlite, shown in Figure~\ref{fg:cover_figure}~(right), is a cheaper version of \locobot that uses the Create2 base~\cite{create2} instead of the Kobuki base.

\textbf{Sawyer}: The Sawyer is a 7-DOF collaborative robot arm from Rethink Robotics \cite{sawyerspec}. \pyrobot interfaces with the Intera SDK provided with the Sawyer.

\textbf{Simulators}: PyRobot currently supports Gazebo simulator~\cite{koenig2004design}, a 3D rigid body simulator popular in the robotics community. For \locobot and \locobotlite, \pyrobot supports tight integration with Gazebo \ie, the same code can be run on both Gazebo and the real robot.


\begin{figure}[t]
  \begin{center}
    \includegraphics[width=1.00\linewidth]{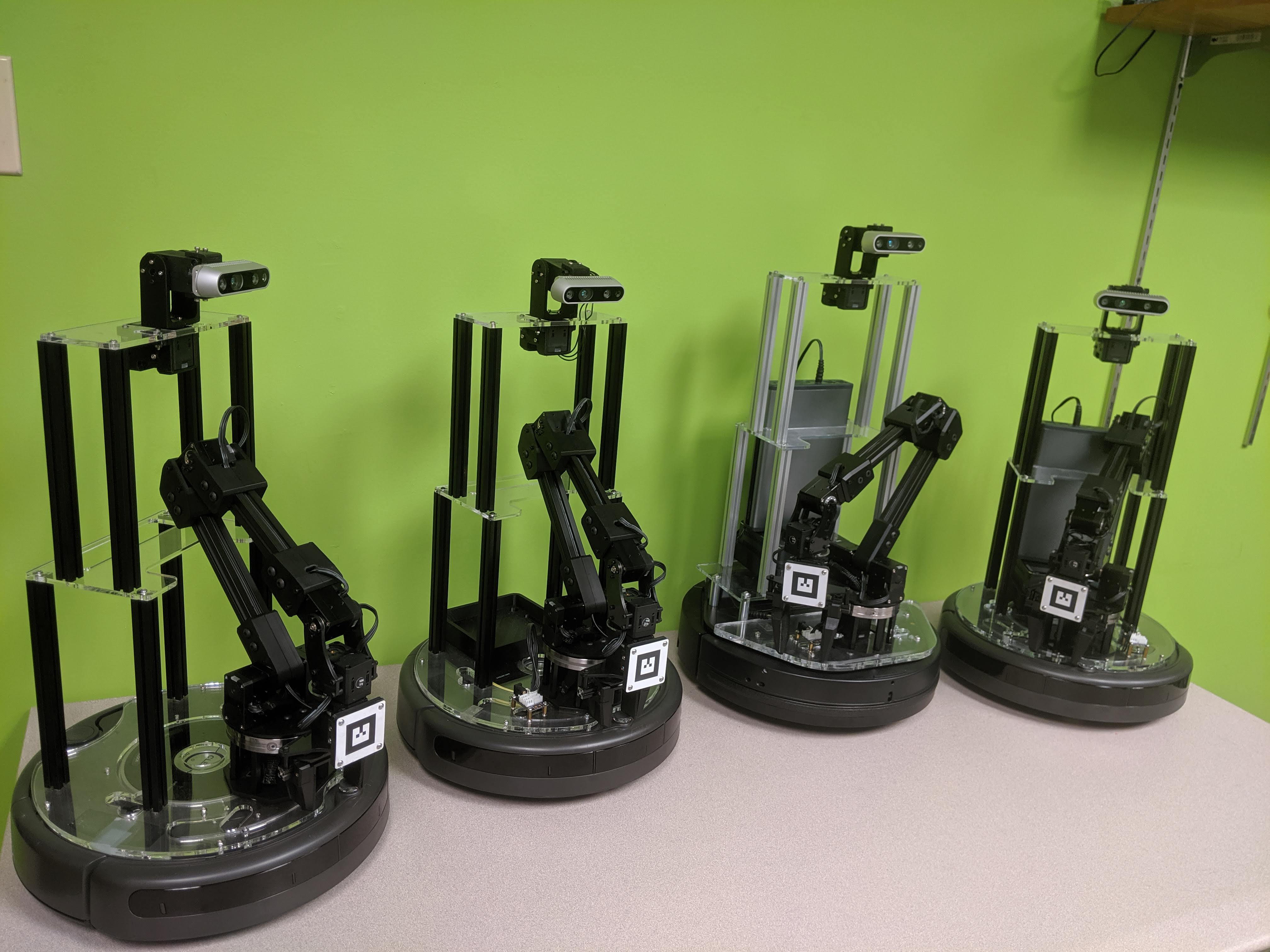} 
    \caption{{\locobot} is low-cost and hence scalable.}
  \label{fg:many}
  \end{center}
\end{figure}

\begin{table*}
\renewcommand{\arraystretch}{1.1}
\setlength{\tabcolsep}{10pt}
\footnotesize
\centering
\caption{\textbf{Base position control performance for \locobot and \locobotlite}. We report translation and rotation error for different
motion types for the different controllers for base position control
implemented in \pyrobot. Lower errors are better.}
\label{tb:locobot_base_error}
\resizebox{1.0\linewidth}{!}{
\begin{tabular}{lcccccccc}
\toprule
 \textbf{ }& \multicolumn{3}{c}{\textbf{Error with respect to motion capture}} & \multicolumn{3}{c}{\textbf{Error with respect to odometry}} \\
 \textbf{Controllers} & ILQR  & Proportional & Movebase & ILQR & Proportional & Movebase\\
\midrule
\multicolumn{7}{l}{\textbf{\locobot}} \\
\quad \textbf{Linear motion}\\
\quad \quad Translation (mm)                & 17 $\pm$ 5 & 46 $\pm$ 23 & 89 $\pm$  16 & 3 $\pm$ 1 & 41 $\pm$ 32 & 102 $\pm$ 2 \\
\quad \quad Rotation (deg)                & 0.43 $\pm$ 0.25  & 1.77 $\pm$ 1.46 & 10.81 $\pm$ 2.19 & 0.12 $\pm$ 0.10 & 1.65 $\pm$ 1.37 & 10.63 $\pm$ 2.19  \\
\quad \textbf{Rotation motion}\\
\quad \quad Translation (mm)                & 6 $\pm$ 0  & 6 $\pm$ 4 & 4 $\pm$ 2 & 0 $\pm$ 0 & 5 $\pm$ \ 1 & 2 $\pm$ 1 \\
\quad \quad Rotation (deg)                &  1.32 $\pm$ 0.68  & 2.48 $\pm$ 0.98 & 12.53 $\pm$ 1.09 & 1.45 $\pm$ 0.24 & 2.54 $\pm$ 1.02 & 13.08 $\pm$ 1.18 \\
\multicolumn{4}{l}{\quad \textbf{Combined motion}}\\
\quad \quad Translation (mm)                & 16 $\pm$ 2 & 65 $\pm$ 52 & 78 $\pm$ 2 & 6 $\pm$ 1 & 55 $\pm$ 50 & 87 $\pm$ 15 \\
\quad \quad Rotation (deg)                & 0.29 $\pm$ 0.19  &  3.2 $\pm$ 2.69 & 11.59 $\pm$ 1.3 & 0.84 $\pm$ 0.20 & 2.35 $\pm$ 2.94 & 11.65 $\pm$ 1.63  \\
\midrule
\multicolumn{7}{l}{\textbf{\locobotlite}} \\
\quad \textbf{Linear motion}\\
\quad \quad Translation (mm)                & 144 $\pm$ 8 & 142 $\pm$ 7 & 260 $\pm$  81 & 9 $\pm$ 5 & 34 $\pm$ 5 & 99 $\pm$ 31\\
\quad \quad Rotation (deg)                & 1.79 $\pm$ 1.59  & 2.82 $\pm$ 0.52 & 7.34 $\pm$ 8.19  & 1.6 $\pm$ 1.5  & 1.61 $\pm$ 0.34  & 5.21 $\pm$ 3.13  \\
\quad \textbf{Rotation motion}\\
\quad \quad Translation (mm)                & 3 $\pm$ 2  & 3 $\pm$ 2 & 3 $\pm$ 1 &2 $\pm$ 2 & 3 $\pm$ 3& 3 $\pm$ 1 \\
\quad \quad Rotation (deg)                &  6.97 $\pm$ 1.71  & 3.07 $\pm$ 3.47 & 9.94 $\pm$ 1.46 & 1.44 $\pm$ 1.12  & 4.59 $\pm$ 2.78  & 3.42 $\pm$ 1.66   \\
\multicolumn{4}{l}{\quad \textbf{Combined motion}}\\
\quad \quad Translation (mm)                & 123 $\pm$ 7 & 99 $\pm$ 4 & 230 $\pm$ 57 & 5 $\pm$ 6 & 93 $\pm$ 19 & 93 $\pm$ 21 \\
\quad \quad Rotation (deg)                & 2.8 $\pm$ 1.68  &  1.19  $\pm$  0.95 & 5.87 $\pm$ 8.22 & 2.57 $\pm$ 1.31  & 1.57 $\pm$ 1.15  & 4.18 $\pm$ 3.45   \\
\bottomrule
\end{tabular}}
\end{table*}


\section{\pyrobot Controllers}
While a number of robots come with their own implementations for low-level control, \pyrobot implements basic controllers for differential drive bases. It also interfaces with planners such as MoveIt!~\cite{chitta2012moveit} and Movebase~\cite{movebase}. We measure the performance of these controllers and planners implemented in PyRobot for the LoCoBot base and arm.

\subsection{Accuracy of Base Control}
\pyrobot implements position controllers to command 
the robot base to a desired target position (parameterized as a 
3-DOF pose, $(x,y)$ location of the base and its heading 
$\theta$: $[x, y, \theta]$). We implement the following three controllers: \\
\textbf{DWA Controller from Movebase}: We implemented Dynamic Window Approach Controller (DWA) \cite{dieterDWA} for our robot through Movebase \cite{movebase} navigation engine. In this approach, we repeatedly sample a discrete sequence in the robot's control space with the highest score and execute the sequence until the target is reached. \\
\textbf{Proportional Controller}: We decompose the motion into an on-spot rotation, linear motion and a final on-spot rotation at the target location. Each segment of this motion is executed using a proportional controller that applies velocities proportional to the tracking error. For smooth motion, we bound the velocities and the change in velocities. \\
\textbf{Linear Quadratic Regulator}: We analytically compute a trajectory (a sharp one that breaks the motion into on-spot rotation, straight motion and a final on-spot rotation; or a smooth one by fitting a b\'ezier curve between the stating state and the ending state). We sample this trajectory to obtain a state trajectory using constraints on maximum linear and angular velocities. We linearize the dynamics of the robot (assumed to be a bicycle model~\cite{aastrom2012introduction}) around this state trajectory, and construct a LQR feedback controller~\cite{aastrom2012introduction} to track this state trajectory.

We conducted trials on the robot to quantify the accuracy of each of these different position controllers on both \locobot and \locobotlite. We measured accuracy using the difference in commanded state \vs the achieved state as measured using a Vicon motion capture system. The error was factored into  translation (difference in $(x,y)$  location), and  rotation (difference in the heading $\theta$). We report these errors in Table \ref{tb:locobot_base_error}. We group trials into the following three categories: \textit{a) Linear motion}: 5 trials each with targets \SI{2}{\m} in front ($[2,0,0]$), or \SI{2}{\m} behind ($[-2,0,0]$); \textit{b) On-spot rotation}: 5 trials each with target being left rotation by $\pi/2$ ($[0, 0, \pi/2]$), right rotation by $\pi/2$ ($[0,0,-\pi/2]$); \textit{c) Combined linear and rotation motion}: 5 trials each with targets $[1,1,0]$ and $[-1,-1,0]$. 

Table~\ref{tb:locobot_base_error} reports translation and rotation errors for the different controllers for the two robots for these different cases. We generally note that errors are lower for \locobot \vs \locobotlite. Additionally, LQR and proportional controller generally perform better than the DWA controller from Movebase. As all these controllers close the loop on the base odometry, we additionally include errors with respect to base odometry in right part of the table. We observe that the LQR controller is more effective
at closing the loop.

\begin{figure}
\centering
\includegraphics[width=1.0\linewidth]{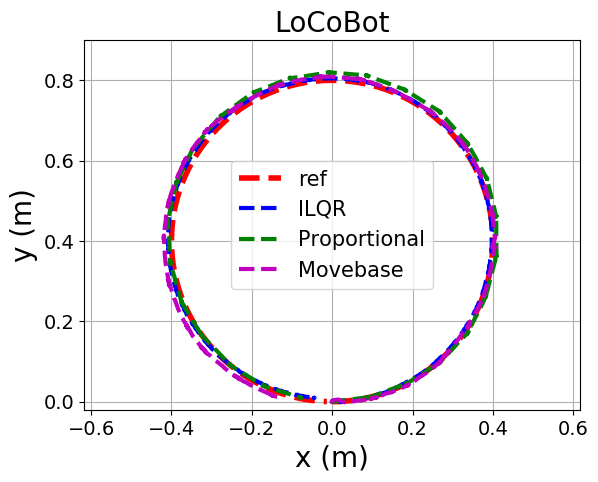}
\includegraphics[width=1.0\linewidth]{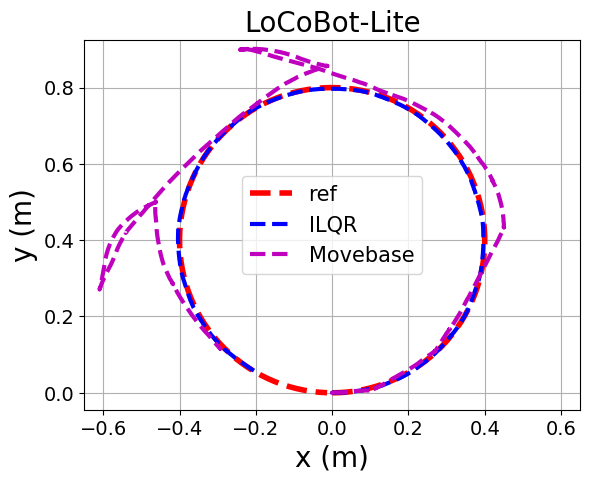}
\caption{Qualitative comparisons for trajectory tacking for \locobot and \locobotlite. Reference trajectory (a circle of radius \SI{0.4}{m}) is shown in red.}
\label{fg:tracking}
\end{figure}

\pyrobot also implements trajectory tracking (using 
feedback controllers as described above). We show qualitative comparisons between different controllers in Figure~\ref{fg:tracking}.

\subsection{Repeatability Tests for Manipulator}
Compared to expensive industrial and collaborative robots, low-cost manipulators like \locobot suffer from control errors that can be attributed to a range of factors: manufacturing and assembling error, gear backlash, hardware execution error, kinematics inaccuracy, hand-eye calibration error, motor wear and tear, etc. The position-control repeatability was analyzed by commanding the arm to 4 different 3D poses (and the home pose) in a 2D grid at a fixed height without carrying a payload for a total of 10 repetitions per pose. The ground truth positions were measured using a Vicon motion capture system at \SI{120}{\hertz}. The arm always started at the home pose (when the joint angles are all 0) before moving to the commanded end pose. The results are summarized in Table \ref{tb:arm_error}. Overall, the arm had a repeatability error of \SI{0.33}{\milli\meter} to \SI{0.58}{\milli\meter}, computed based on ISO9283 standard. Poses 1 and 3 were closer to the robot torso and had lower error compared to Pose 2 and 4 where the arms were extended at the extremities of the workspace. The standard deviation along the z axis was also higher across all poses due to gravity. For comparison, the Sawyer and UR5 robots are reported to have a repeatability of \SI{0.1}{\milli\meter} \cite{ur5spec, sawyerspec}. The position control in the initial release only relies on proprioceptive feedback, and using feedforward model-based control in future release could reduce the error further. The PID gain settings are exposed to the user for more specialized robot or task-specific tuning.

\begin{table}
\footnotesize
\centering
\caption{Locobot Arm Pose Repeatability}
\label{tb:arm_error}
\begin{tabular}{lccccc}
\toprule
\multirow{2}{*}{\textbf{Std Dev.(mm)}} & \multicolumn{5}{c}{\textbf{Poses}} \\
  & 1  & 2 & 3 & 4 & Home \\
\midrule
x                & 0.12 & 0.13 & 0.07 & 0.11 & 0.15  \\
y                & 0.13 & 0.07 & 0.10 & 0.14 & 0.27   \\
z                & 0.21 & 0.33 & 0.22 & 0.31 & 0.24  \\
\midrule
\textbf{Repeatability (mm)}       & 0.41 & 0.58 & 0.33 & 0.50 & 0.52      \\
\bottomrule
\end{tabular}
\end{table}

\section{High-Level AI Applications}
We discuss implementation of a few example high-level AI applications through the \pyrobot API.

\subsection{Visual SLAM}\label{sec:slam}
Visual SLAM algorithms provide more accurate odometry as compared to odometry that is derived purely from inertial sensors on the base. We deployed ORB-SLAM2~\cite{murORB2}, a leading visual SLAM systems in the \pyrobot library. ORB-SLAM2 is a feature-based indirect visual SLAM system that uses ORB features to perform tracking, mapping, and loop closing. We adapt the open-source ORB-SLAM2 code into a ROS package. This package saves RGB and depth images of the keyframes and continuously publishes camera trajectory and camera pose. PyRobot uses this published pose information to return the robot base state and trajectory. This state derived from visual SLAM can be used in downstream controllers or algorithms for more accurate behavior. \pyrobot also supports dense map reconstruction, by integrating depth image observations using the ORB-SLAM2 estimated camera pose. This can be used for motion planning for navigation tasks.
\begin{figure*}[h!]
\centering
\includegraphics[width=1.0\linewidth]{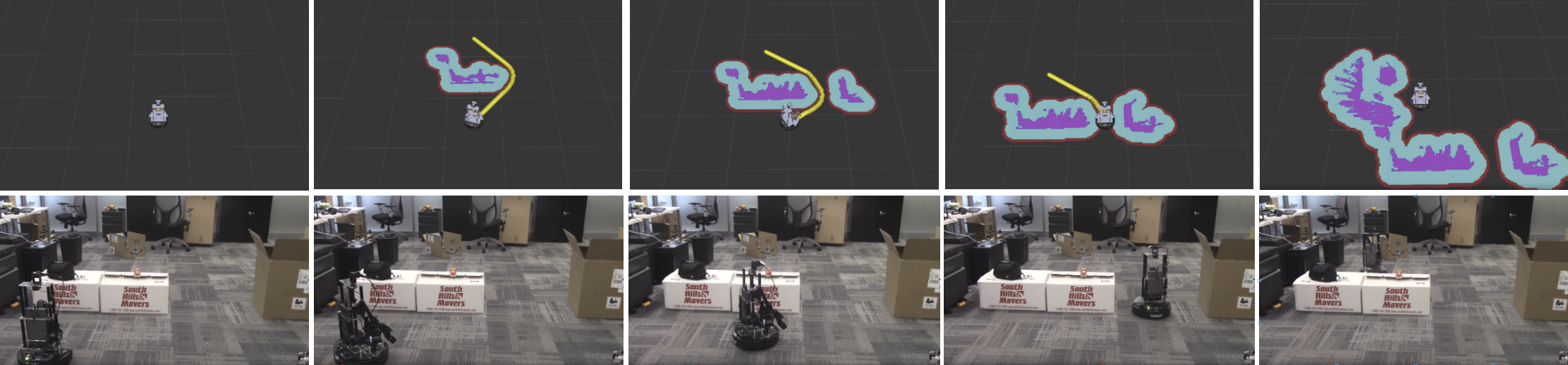}
\caption{An example of Navigation via SLAM and Path Planning. First row corresponds to the 2-D map constructed using the on-board SLAM and the second row corresponds to the actual motion of the robot.}
\label{fg:mapping}
\end{figure*}

\subsection{Navigation via SLAM and Path Planning}
We deployed Movebase~\cite{movebase} ROS package on LoCoBot and \locobotlite for safe navigation in environments with obstacles. We use the occupancy map as obtained from visual SLAM, to compute a 2D cost-map that denotes regions of the environment where the robot is safe to move. Movebase uses this cost-map to generate collision free trajectories to goals specified in the environment. These trajectories can be executed using any of the controllers implemented in \pyrobot. These steps are run continuously, and the plan is updated if it becomes infeasible as the robot perceives previously unseen parts of the environment.

\subsection{Learned Visual Navigation}
\begin{figure*}
\centering
\includegraphics[width=0.195\linewidth]{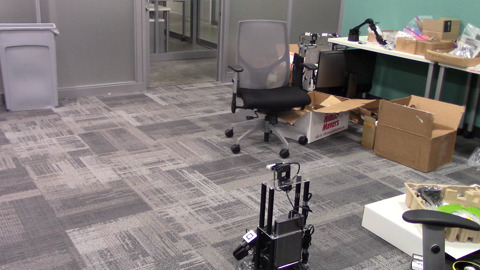}
\includegraphics[width=0.195\linewidth]{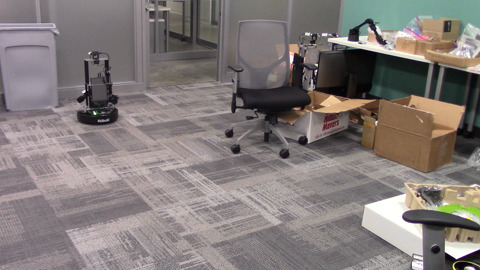}
\includegraphics[width=0.195\linewidth]{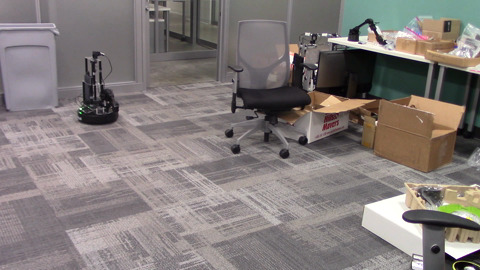}
\includegraphics[width=0.195\linewidth]{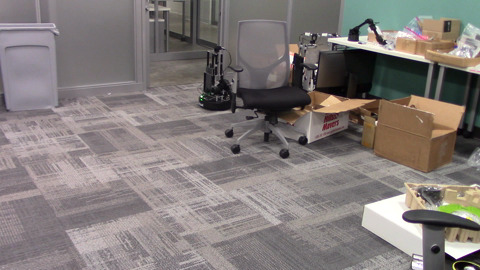}
\includegraphics[width=0.195\linewidth]{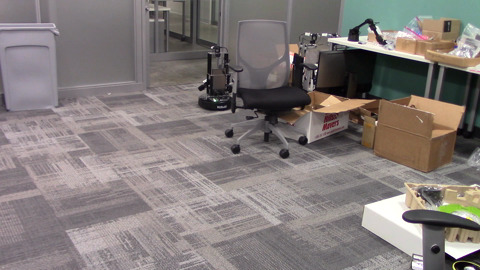}
\caption{Snapshots from a run of visual navigation policy (CMP~\cite{gupta2017cognitive}) deployed on \locobot. See project website for videos.} 
\label{fg:cmp}
\end{figure*}

We deploy learned policies for visual navigation on \locobot
using \pyrobot API. We work with the cognitive mapping and planning
policy (CMP) from Gupta \etal~\cite{gupta2017cognitive}. Given an input
goal location, CMP policy takes in the current image from the on-board 
camera to output one of four macro-actions 
(stop, turn left, turn right or go straight). We use the base position 
control interface in \pyrobot API to execute these actions. 
Listing~\ref{code:cmp} shows simplified code, and Figure~\ref{fg:cmp} shows
frames from a sample execution.

\subsection{Grasping}

\begin{figure*}[h!]
  \begin{center}
    \includegraphics[width = 1.0\linewidth]{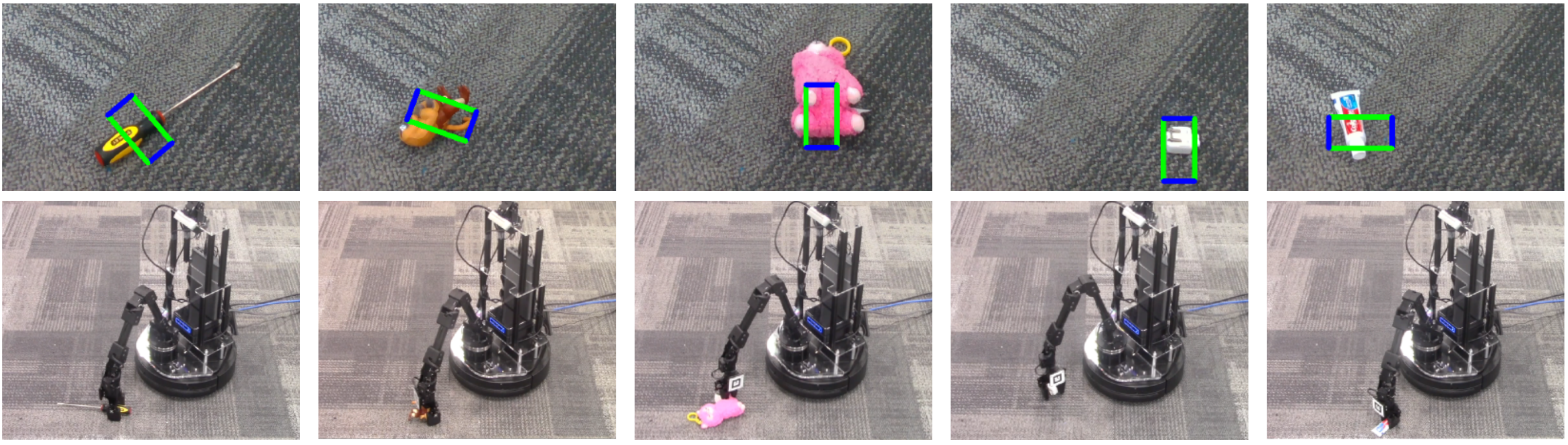}
    \caption{Grasps selected by the grasp model and execution by the robot.
}
  \label{fg:grasps}
  \end{center}
\end{figure*}

We deploy a learned-based grasping algorithm to grasp objects placed on the ground from RGB images using the PyRobot API. The model is trained on data from people's homes~\cite{robotsinhome} and is robust to a wide variety of objects and backgrounds. This model outputs a grasp in the image space. This grasp is parameterized by 2D location in the image and the gripper orientation. We convert this 2D location and orientation into the \textit{grasp position} (3D location and orientation) using known camera parameters, and the depth image. We command the robot to the \textit{pre-grasp location}, that is a few centimeter above the grasp position, lower the arm to reach the object, and close the gripper to grasp the object.
Listing~\ref{code:grasping} shows simplified code, and Figure~\ref{fg:grasps} shows sample grasps using the LoCoBot.

\subsection{Pushing}
We deploy a heuristic-based pushing algorithm using \pyrobot. It relies on the depth sensor, and thus the quality of the pushing depends on how well the stereo-based depth sensor behaves in different background. To achieve the best performance, it is best to place the robot on a floor with non-uniform texture. 

The algorithm can be summarized with the following steps: (1) Move the arm out of the camera's field of view. (2) Filter the point cloud seen by the RGBD camera, specifically removing points too far away and those that correspond to the floor by coordinate thresholding. (3) Project the remaining point cloud onto the xy-plane and use DBSCAN \cite{ester1996density} algorithm to automatically cluster the projected points. (4) Randomly select one cluster and choose a random push-start point on the enclosing bounding box of the cluster. (5) Move the gripper to the push-start point and move the gripper horizontally towards the center of the cluster. Listing~\ref{code:pushing} shows simplified code.

\section{Related Work}

\textbf{Robotics Software Design.}
The robotics community has embraced a layered hierarchical software design from the early days \cite{brooks} and re-usability has been a core design principle \cite{thinsoftware}. We refer readers to Tsardoulias and Mitkas~\cite{emmanouil2017} for a comprehensive review. There have been several motion planning libraries such as OpenRave \cite{diankov08}, MoveIt! \cite{chitta2012moveit}, OMPL \cite{ompl} which provide hardware-agnostic core functionalities that can be compiled for each specific robot. In the likes of ROS, there have also been robotics ecosystems, such as OROCOS \cite{herman2001} and the Microsoft Robotics Studio that support kinematic libraries, distributed processes, state machines for the real time control of robots.

\textbf{Low-cost Mobile Manipulators.}
There has been very limited research on learning on low-cost robots, given that most researchers use standard industrial or collaborative robots. Deisenroth \etal~\cite{deisenroth2011} used model-based RL to teach a cheap inaccurate 6 DOF robot to stack multiple blocks and a previous iteration of LoCoBot was used in Gupta \etal~\cite{robotsinhome} to learn visual grasping policies with real data collected in people's homes. Recently, Gealy \etal~\cite{blue2019} proposed a compliant low-cost arm using quasi-direct drive actuation.

\textbf{Open Source Manipulators.}
There has been very limited work in open sourced manipulators. Raven is a open architecture surgical research robot \cite{raven}. Recently, the Open Manipulator project from Robotis allows one to build their own low cost robot with custom kinematics and design \cite{robotis}.

\textbf{Research Ecosystems in AI Fields.} Research in a number of AI fields has benefited from there being common tasks (such as object detection in computer vision or parsing in NLP), common datasets (such as BSDS~\cite{MartinFTM01}, ImageNet~\cite{ILSVRC15}, PASCAL VOC~\cite{Everingham15} and MSCOCO~\cite{lin2014microsoft} in computer vision, or Penn Tree Bank~\cite{marcus1993building}, GLUE~\cite{wang2019glue}, SentEval~\cite{conneau2018senteval} and WMT in NLP, \etc), and common code bases to experiment with (DPMs~\cite{voc-release5}, Caffe~\cite{jia2014caffe}, Stanford CoreNLP~\cite{manning2014stanford}, spaCy~\cite{spacy2}, \etc). While some people argue that such use of common tasks and datasets can prevent creative progress, at the same time, it has lead to rapid progress in these fields, as researchers can quickly replicate results and build upon each other work.

\textbf{Benchmarking in Robotics.} Benchmarking in robotics is extremely challenging given the vast scope of applications and diversity of physical test conditions (hardware, objects, environment, etc.). It is a well acknowledged concern within the robotics community that we are yet to develop reliable benchmarking metrics that can be widely adopted to quantify research progress. Several workshops have tried to stimulate discourse towards this end \cite{icraworkshop, icraworkshopSLAM} and different task specific metrics have been proposed for grasping \cite{mahler2018}, gripper design \cite{grippermetric}, SLAM \cite{icraworkshopSLAM}, etc. Research has also benefited from creating object datasets with shape and grasp information, such as the Columbia Grasp Database \cite{columbiaGDB}, DexNet \cite{mahler2016dexnet} and KIT Object Models \cite{kit2012}, which could be used for perception and motion planning. The YCB dataset went a step further by distributing a physical dataset of household and kitchen objects with corresponding meta data (shape, RGBD scans, etc) \cite{ycbdataset}. While there is no consensus yet on benchmarking in robotics, we hope that the combination of PyRobot and LoCoBot will facilitate further discussion.
\section{Discussion}
In this paper, we describe the \pyrobot framework, which provides a high-level hardware independent API to control different robots. We believe \pyrobot when combined with low-cost robots such as {\locobot}, will reduce the barrier to entry into robotics. In the immediate future, we will continue to grow the functionality in \pyrobot such as by interfacing with simulators (like AI Habitat~\cite{habitat19arxiv}, Gibson~\cite{xiazamirhe2018gibsonenv} and MuJoCo~\cite{todorov2012mujoco}), improving controllers such as be implementing gravity compensation for \locobot. But more broadly, we believe \pyrobot will lead to the development of a research and teaching ecosystem.


\textbf{\pyrobot for robotics instruction.} Having a beginner-friendly and open architecture is great for robotics education, as affordable robotic setups with LoCoBot and PyRobot could easily be assembled and scaled for hands-on instruction. 10 LoCoBots were used in the Spring 2019 offering of 16-662: Robot Autonomy (by Professor Oliver Kroemer) in the Robotics Institute at CMU, to support homework assignments and projects. 
We believe many more such courses will follow.

\textbf{\pyrobot as a research ecosystem.} Compared to other fields, benchmarking in robotics is challenging due to several reasons. \pyrobot's unified API and \locobot's standard hardware, will allow researchers to share their high level algorithmic implementations, models and datasets collected on a real robot. This will allow researchers to collaborate and iterate faster on robotics applications. We will continue to expand the set of pre-trained models. Hopefully, other researchers will find the \pyrobot framework useful and contribute their models for others to use as well.




\section{Acknowledgements}
We would like to thank Soumith Chintala for countless discussions and providing software engineering guidance. We would also like to thank Deepak Pathak and Shubham Tulsiani for testing, advice and discussions. Finally, we would like to thank Oliver Kroemer, Timothy Lee and Mohit Sharma for introducing LoCoBots in teaching \textit{16-662: Robot Autonomy} at CMU, and Justin MacEy for helping with the motion capture experiments.



\bibliographystyle{ieee}
\bibliography{references}

\appendix
\pagebreak
\section{Code Listings}
\begin{listing}
\caption{Visual navigation example using \pyrobot API.}
\label{code:cmp}
\begin{minted}{python}
from pyrobot import Robot

# Construct Robot.
bot = Robot('locobot')

# Construct policy.
policy = CMP()

# Relative position for each action.
dv = 0.4        # Forward step size
dw = np.pi/2.   # Rotation step size
action_position = [[0., 0., 0.0],
                   [0., 0., -dw],
                   [0., 0., +dw],
                   [dv, 0., 0.0]]

# Set goal for policy.
policy.set_new_goal(goal)
while action != 0:
    # Get image.
    rgb = bot.camera.get_rgb()
    
    # Compute action.
    action = policy.compute_action(rgb)
    
    # Execute action.
    position = action_position[action]
    bot.base.go_to_relative(position)
\end{minted}
\end{listing}

\begin{listing}
\caption{Grasping example using \pyrobot API.}
\label{code:grasping}
\begin{minted}{python}
from pyrobot import Robot

# Construct Robot.
bot = Robot('locobot')

# Set pregrasp and grasp height.
pregrasp_height = 0.2
grasp_height = 0.13

# Construct grasp model.
model = GraspModel()

# Move arm and camera to reset position.
reset_pos = [-1.5, 0.5, 0.3, -0.7, 0.]
bot.arm.set_joint_positions(reset_pos)
bot.camera.set_pan_tilt(0.0, 0.8)

# Get image.
rgb = bot.camera.get_rgb()

# Compute action.
grasp_img = model.compute_grasp(rgb)

# Convert grasp from Image space to 
# robot workspace.
grasp_pose = cvt_space(grasp_img)

# Execute grasp.
# 1. Go to pre-grasp pose
pregrasp_position = [grasp_pose[0], 
                     grasp_pose[1], 
                     pregrasp_height]
grasp_angle = grasp_pose[2]
bot.arm.set_ee_pose_pitch_roll(
    position=pregrasp_position,
    pitch=np.pi / 2,
    roll=grasp_angle,
    plan=False,
    numerical=False)

# 2. Go to grasp pose.
grasp_position = [grasp_pose[0], 
                  grasp_pose[1],
                  grasp_height]
bot.arm.set_ee_pose_pitch_roll(
    position=grasp_position,
    pitch=np.pi / 2,
    roll=grasp_angle,
    plan=False,
    numerical=False)

# 3. Grasp the object
bot.gripper.close()
\end{minted}
\end{listing}
\begin{listing}
\caption{Object pushing example using \pyrobot API.}
\label{code:pushing}
\begin{minted}{python}
from pyrobot import Robot

# Construct Robot.
bot = Robot('locobot')

# Setup gripper, camera, arm.
bot.gripper.close()
bot.camera.set_pan_tilt(0, 0.7, wait=True)

# Move hand out of camera view.
ov_pos = [1.96, 0.52, -0.51, 1.67, 0.01]
bot.arm.set_joint_positions(ov_pos, plan=False)

# Get the point cloud(in base frame).
pts, colors = bot.camera.get_current_pcd(
                  in_cam=False)

# Compute push location, direction.
pre_push_pt, push_pt, obj_center = \
    get_push_direction(pts, colors)

# Move the gripper to pre-pushing pose
bot.arm.set_ee_pose_pitch_roll(
    position=pre_push_pt,
    pitch=np.pi / 2,
    roll=0,
    plan=False,
    numerical=False)
                               
# Move the gripper vertically down.
down_disp = push_pt - pre_push_pt
bot.arm.move_ee_xyz(down_disp, 
                    plan=False, 
                    numerical=False)

# Move the gripper horizontally
# to push the object.
hor_disp = 2 * (obj_center - push_pt)
bot.arm.move_ee_xyz(hor_disp, 
                    plan=False, 
                    numerical=False)
\end{minted}
\end{listing}

\end{document}